\documentclass[10pt,letterpaper]{article}

\usepackage{cogsci}
\usepackage{hyperref}
\usepackage{graphicx}
\usepackage{graphics}
\cogscifinalcopy 
\usepackage{multirow}
\usepackage[T2A, T1]{fontenc}
\usepackage[utf8]{inputenc}
\usepackage[russian, english]{babel}

\usepackage{pslatex}
\usepackage{apacite}
\usepackage{float} 

\title{Falling Through the Gaps: Neural Architectures as Models of Morphological Rule Learning}
\author{
 {\large \bf Deniz Beser (beser@isi.edu)} \\
 Information Sciences Institute \\
 University of Southern California \\
 Marina del Rey, CA 90292
 }
\begin{document}

\maketitle
\begin{abstract}
Recent advances in neural architectures have revived the problem of morphological rule learning. We evaluate the Transformer as a model of morphological rule learning and compare it with Recurrent Neural Networks (RNN) on English, German, and Russian. We bring to the fore a hitherto overlooked problem, the morphological gaps, where the expected inflection of a word is missing. For example, 63 Russian verbs lack a first-person-singular present form such that one cannot comfortably say \textit{*oščušču} (`I feel'). Even English has gaps, such as the past participle of \textit{stride}: the function of morphological inflection can be partial. Both neural architectures produce inflections that ought to be missing. Analyses reveal that Transformers recapitulate the statistical distribution of inflections in the training data, similar to RNNs. Models' success on English and German is driven by the fact that rules in these languages can be identified with the majority forms, which is not universal.\footnote{Our code and data are available at 
\href{https://github.com/denizbeser/gaps}{github.com/denizbeser/gaps}}


\end{abstract}

\section{Introduction}
A well-recognized aspect of language acquisition is the learning of productive morphological rules, such as the English past tense (e.g. \textit{walk + /-d/; walked}) \cite{berko1958child, albright2003rules, lignos2018morphology}. Neural networks were first proposed as models of learning such morphological rules decades ago \cite{rumelhart1985learning}, and were faced with a seminal critique that highlighted their numerous linguistic inadequacies \cite{pinker1988language}. In light of the recent advancements in the field of natural language processing (NLP), the encoder-decoder recurrent neural network (RNN) architecture, a sequence-to-sequence model commonly used for machine translation \cite{sutskeverRNN}, has been reconsidered as a model for learning productive morphological rules \cite{kirov2018recurrent}.  \citeA{corkery2019we} compared the performance of the RNN architecture to human behavior on English past tense, and concluded that the fit to human data is weak. Similarly, \citeA{mccurdy2020inflecting} used the German plural to demonstrate that the RNN's behavior does not match human subjects and the model is susceptible to frequency distribution in the data, an argument first raised by \citeA{marcus1995german}. 

Evidently, the encoder-decoder RNN is insufficent as a cognitive model of morphological rule learning. However, RNNs are no longer regarded as the state-of-the-art system for language modeling in NLP. \citeA{vaswani2017attention} introduced the Transformer architecture, a neural architecture that relies on attention mechanism and does not use recurrence. The Transformer architecture has since outperformed RNNs across a wide range of tasks \cite{vaswani2017attention, liu2019roberta, brown2020language}, including morphological inflection generation \cite{wu2020applying}. The outstanding performance of the Transformer calls for an evaluation as a cognitive model for learning productive morphological rules.

In this paper, we first compare the performances of the Transformer architecture and the encoder-decoder RNNs across morphological phenomena in natural language that suggest humans seek certain structures in the languages they hear in order to learn productive morphological rules. Namely, we use the productivity observed in English past tense \cite{berko1958child, albright2003rules} and the non-majority productivity in German plurals \cite{elsen2002acquisition,bartke2005erp,zaretsky2015no} which have previously been used for assessing RNNs as models of morphological learning \cite{kirov2018recurrent, corkery2019we, mccurdy2020conditioning}. We then evaluate both architectures on a set of Russian verbs, an example of gaps in morphological productivity where native speakers of the language fail to generalize a productive rule and produce an inflected form \cite{sims2006minding,pertsova2016transderivational,gorman2017nobody}. Our analyses on the English and German languages show that the Transformer architecture is susceptible to the same analogical and distributional errors as RNNs. Likewise, both neural architectures produce inflected forms for the gapped Russian verbs while native speakers cannot. We argue that the models' predictions on the gapped words evince a fundamental challenge for both Transformers and RNNs as models of language learning and underline the need for additional structural priors for neural models to capture how humans expect and utilize structure in language data to generalize a rule productively.

\section{Background}
\subsection{Majority Productivity: The English Past Tense}
What is the past tense of the verb \textit{google}? The English past tense is a prevalent example of morphological productivity in linguistic literature; a vast majority of English verbs follow a typical inflectional pattern (e.g. \textit{walk-walked, kiss-kissed}) and a smaller subset of the verbs are exceptions (e.g. \textit{run-ran, think-thought}) \cite{rumelhart1985learning, xu1995weird, albright2003rules, lignos2018morphology}. Children can learn these morphological rules from a young age despite sparsity of the data and even though they might not have heard the past tense of \textit{`look'} before, they can effortlessly produce the past tense \textit{`looked'} \cite{berko1958child}. They do not memorize all the words and their past tenses, but extract a productive rule and apply it to novel words. Importantly, this productive rule-learning is categorical in nature as children consistently apply to novel words and generally do not make analogical errors of `overirregularization'; they categorically produce the past tense pair \textit{bling-blinged}, despite the existence of an irregular pair \textit{sing-sang}) \cite{berko1958child,macwhinney1978acquisition,xu1995weird}. This rule-learning paradigm is further evindenced by patterns in language acquisition, including overgeneralization errors \cite{marcus1992overregularization} as children tend to apply learned rules to irregular verbs to produce regular inflections (e.g. \textit{think-thinked}) and then correct the behavior over time. Previous work on the RNN's effectiveness on modeling the English past tense found that the model performs worse than rule-based systems and does not correlate strongly with human behavior \cite{corkery2019we}. 

Since most of the English verbs follow this productive rule, a simple majority-rule could explain the productivity of the English past tense; however, there are telltale paradigms observed in language acquisition across different languages which show that the learning behavior is much more complicated, where the majority rule is not the productive one \cite{zaretsky2015no,gorman2017nobody}.

\subsection{Non-Majority Productivity: The German Plural}
\begin{figure}[h!]
  \centering
  \includegraphics[scale=0.275]{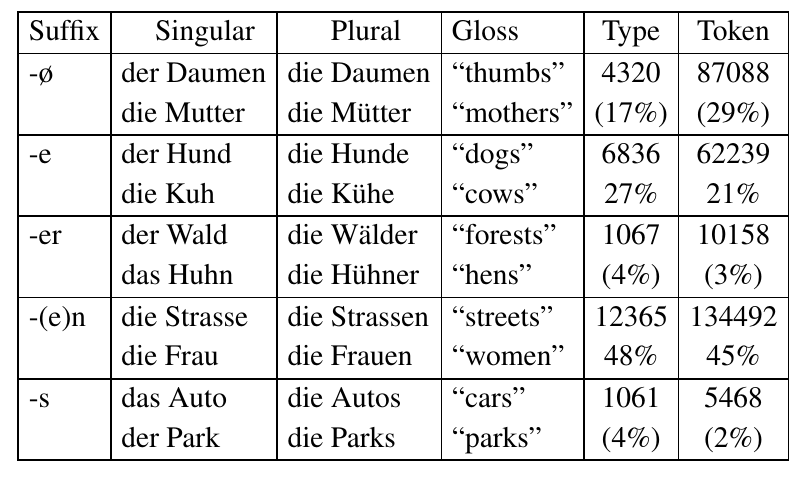}
  \caption{Types of German plural suffixes and their distribution with examples, based on the CELEX corpus \protect\cite{sonnenstuhl2002processing,yang2016price}.}
  \label{fig:german-freqs}
\end{figure}

While the English past tense can be expressed by a simple majority rule, sometimes no marker may form a majority, and even the least frequent marker may be applied productively \cite{elsen2002acquisition, bartke2005erp, yang2016price}. The German plural is a fitting example for non-majority productivity in rule learning as native speakers use multiple plural markers productively \cite{zaretsky2015no, mccurdy2020inflecting}. German has three grammatical genders and primarily five different types of suffixes that mark plural nouns (e.g. \textit{die Frau - die Frauen}) partially categorized by gender. The most frequent of these markers is the /-(e)n/ with 48\% type and 45\% token ratio, and the least frequent of these is /-s/ with 4\% type and 2\% token ratio. The other three common suffix types for German plurals are /-e/, /-er/, and /-Ø/ denoting no change \cite{sonnenstuhl2002processing,yang2016price}. 
See figure \ref{fig:german-freqs} for an overview of the distribution of plural markers in German.

German speakers certainly do not memorize the plural form of every noun; despite the lack of an apparent majority, they learn to produce plural forms using productive rules. However, there has been mixed arguments on the nature of the German plural: while the minority plural suffix /-s/ was first found to be applied to novel words \cite{clahsen1992regular} and argued to be the default productive marker \cite{marcus1995german}, recent research suggests that the more frequent suffixes /-e/ and /-(e)n/ are also used productively on novel words \cite{zaretsky2015no} and the inflections are influenced by both gender and phonological properties \cite{mccurdy2020conditioning}. Our analyses in this paper follow the perspective of the most recent studies on this subject that suggest there are multiple productive rules applied based on gender and phonological features \cite{zaretsky2015no,mccurdy2020inflecting}. Previous work on evaluating RNNs performance on German plurals have found that the model is biased toward the most frequent suffix \cite{mccurdy2020inflecting} and is oversensitive to gender \cite{mccurdy2020conditioning}.

\begin{table*}[t]
\centering
\begin{tabular}{lcccccccccc}
\textbf{} & \multicolumn{4}{c}{\textbf{RNN}} & \multicolumn{4}{c|}{\textbf{Transformer}} & \multicolumn{1}{c}{\multirow{2}{*}{\textbf{\begin{tabular}[c]{@{}c@{}}Prediction\\ Correlation\end{tabular}}}} & \multicolumn{1}{c}{\multirow{2}{*}{\textbf{\begin{tabular}[c]{@{}c@{}}Prediction\\ Similarity\end{tabular}}}} \\ \cline{2-9}
 & \textbf{Train}& \textbf{Dev}& \textbf{Test} & \textbf{{/}-d{/} \%} & \textbf{Train}& \textbf{Dev}& \textbf{Test} & \multicolumn{1}{l|}{\textbf{{/}-d{/} \%}} & \multicolumn{1}{c}{} & \multicolumn{1}{c}{} \\
 \hline\hline
\textbf{All Verbs} & 96.8 & 98.2 & 80.2 & \textbf{97.6 }& 99.3 & 98.6 & 94.0 & \textbf{99.6} & 0.877 & 94.6 \\
\textbf{Irregulars} & - & - & 18.5 & \textbf{96.3} & - & -  & 7.4 & \textbf{92.6 }& 0.734 & 88.9
\end{tabular}
\caption{Breakdown of model performance on the two English \textit{test} sets, one that consists of regulars and irregulars, and one that consists of irregulars. The model is trained once on all verbs, a set of regular and irregular verbs, and then tested on both all verbs and a subset of irregular verbs. The {/}-d{/} \% column reports the percentage of predictions that use the {/}-d{/} productive rule. The last two columns compare the Spearman correlation (\textit{p<0.01}) and similarity of models' inflection predictions on the test set as explained in the \hyperref[sec:eval-overview]{Evaluation Overview section}.}
\label{table:english-results-2}
\end{table*}

\subsection{Morphological Gaps: Defective Verbs in Russian}
Besides the majority productivity of English past tense and non-majority rules observed in German plurals, there are numerous cases across languages where there is a gap in productivity, where the speakers fail to comfortably produce an inflected form, despite an apparent majority pattern \cite{sims2006minding, pertsova2016transderivational, gorman2017nobody}. For instance, consider the past participle of the English verb \textit{stride}. Native speakers of English struggle to pick a past participle for this verb (\textit{strode, stridden, strided, ...}) as there is no apparent productive rule for English past participle \cite{gorman2017nobody}. 

In Russian, there is a defective gap of this kind in 63 second conjugation verbs \cite{sims2006minding,pertsova2016transderivational}. These defective verbs have stems ending in dental consonants (\textit{t, d, s, z}), and in the first person singular (1SG) present form, they are expected to alternate with a palatal or alveo-palatal fricative \cite{pertsova2016transderivational}. Yet, native speakers of Russian struggle to come up with an absolute 1SG word form for this set of second conjugation verbs \cite{pertsova2016transderivational}. Some examples of these gapped verbs include \textit{‘to vacuum’ (*pylesosit’)}, \textit{`to feel' (*oščutit’)}, and \textit{`to misbehave' (*škodit’)}. We evaluate how this paradigm may pose a challenge for the Transformer and RNNs due to neural networks' distributional tendencies \cite{corkery2019we, mccurdy2020conditioning}

\section{Evaluations}
\subsubsection{Overview} \label{sec:eval-overview} To compare the performances of the Transformer and RNN architectures on morphological patterns in each language, we train the separate instances of both architectures on pairs of bare and inflected word forms, where each word is represented as a sequence of space-separated characters \cite{kirov2018recurrent, wu2020applying} In other words, the models learn to translate one sequence of characters to another. During testing, we perform `wug tests' on the models by presenting novel words to the models and evaluating the inflected word forms the models predict \cite{berko1958child}. In our evaluations, we assess prediction accuracy on relevant test sets for the language and the distribution of inflections predicted by the models. We also measure the Pearson correlation coefficient and \textit{prediction similarity} (measured as accuracy across predictions) between the inflection predictions of the RNN and Transformer architectures.
\subsubsection{Implementation Details} The Transformer models use eight attention heads, have two layers, model dimension of 312, and feed-forward network dimension of 512. The decoder uses beam search with beam size of 12. The Transformer models are trained for 20,000 steps. 

The RNN implementation is kept identical to \citeA{kirov2018recurrent}. The RNNs are trained for 100 epochs, with a batch size of 20. Both encoder and decoder RNNs are bidirectional LSTMs \cite{schuster1997bidirectional} with two layers and 100 hidden units, with a vector size of 300. The RNN decoder also uses beam search with beam size of 12. Both the RNN and the Transformer implementations are based on OpenNMT \cite{klein2017opennmt}.

\subsection{The English Past Tense}
\subsubsection{Method} We train the models on pairs of English verbs with the infinitive and past-tense form sampled from English CELEX \cite{baayen1995celex2}. The verbs are represented in phonetic form. The dataset uses type frequency for training; i.e each verb appears once in the training dataset. 
The dataset had 4,000 datapoints for training, 500 for development, and 500 for testing. During testing we use test accuracy, the prediction rate of the /-d/ majority productive rule, and the correlation and similarity between the models. In addition to evaluating the model performance on the complete test set which contains both regular and irregular English past tense verbs, we also measure the performance on the irregular subset of test verbs.

\begin{figure*}[t]
  \centering
  \includegraphics[scale=0.35]{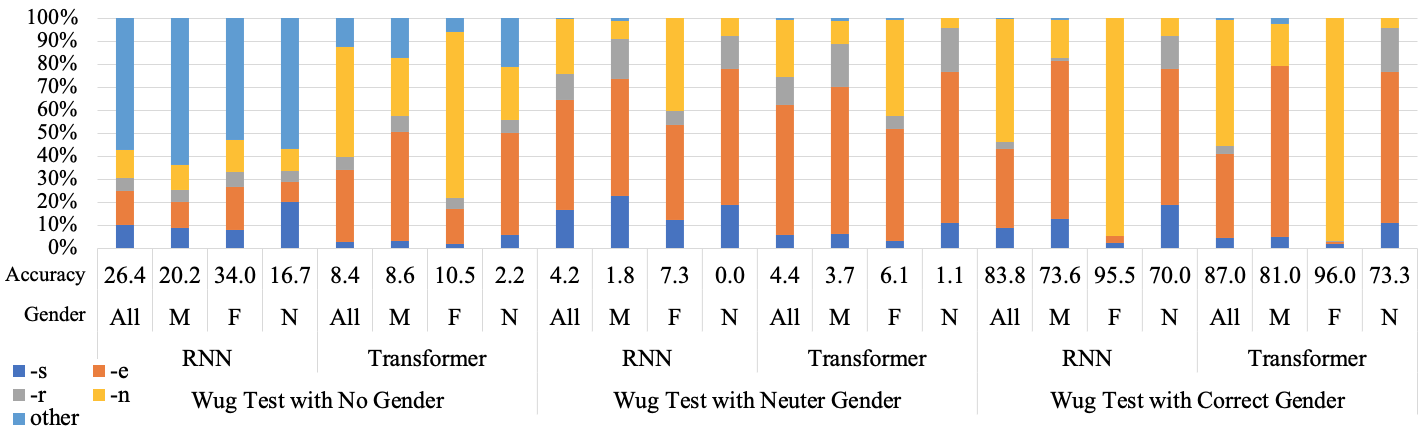}
  \caption{The cumulative distribution of predicted inflections for German nouns across model types and test datasets. Prediction accuracy on test sets separated by gender are marked on the x axis.}
  \label{fig:german-plot}
\end{figure*}

\subsubsection{Results and Discussion}
The results of English evaluation are shown in Table \ref{table:english-results-2}. The Transformer architecture achieves a higher prediction accuracy and /-d/ prediction rate than the RNN when tested on all 500 test verbs. This trend is reversed when the models are tested on the irregular verbs; in this case, the Transformer has a lower test accuracy and /-d/ prediction rate than the RNN. There is a high resemblance between the predictions made by the models; when tested on all verbs, the correlation between the models is 0.877, and 94.6\% of predictions are identical. When tested on irregular verbs, the correlation and the similarity between the models are both lower, 0.734 and 88.9\%, respectively. Both models achieve high /-d/ rule prediction rates, meaning that they prioritize the correct productive rule. However, when this rule prediction rate is compared across test conditions, we see that both models have a lower rate when tested on irregular verbs. The drop in the /-d/ prediction rate when tested on irregulars, which is higher for the Transformer than for the RNN, suggests that both models, especially the Transformer, are susceptible to analogical errors that are made by adults but not by children \cite{berko1958child, albright2003rules} and that neither model does not in fact learn a single productive rule.


\subsection{The German Plural}
\subsubsection{Method 1: Model Comparison} To compare the models on the acquisition of the German plural, we train both networks with singular-plural verb pairs marked with their gender. \footnote{We use the German CELEX dataset \cite{baayen1995celex2}} The training, development, and test datasets for German had 4000, 500, and 500 datapoints, respectively. After training is complete, each model is wug-tested on the two test sets. In the first dataset, all test nouns are marked with neuter gender. In the second dataset, each noun is marked with its correct gender. We then evaluate the prediction rates of each plural marker on different subsets of the test datasets, separated by the original gender marker of the nouns in the dataset.

\subsubsection{Method 2: Human Data Comparison}
In our second evaluation for German, we use data provided by \citeA{mccurdy2020inflecting} to compare the two models with the human behavior. Similar to \citeA{mccurdy2020inflecting}, we train 25 separate instances of the RNN and Transformers. For each model instance, we use a different 4000/500/500 train/dev/test random split from 5000 German CELEX verbs pairs in orthographic form.\footnote{The RNN is trained for 50 epochs and the Transformer is trained for 10,000 steps.} We test the models on the CELEX test set, and use the experiment test set provided by \citeA{mccurdy2020inflecting} for human behavior comparison.

\subsubsection{Results and Discussion}
The test performance for both models with the distribution of the predictions for each test set are shown in Figure \ref{fig:german-plot}. The models achieve a high test accuracy and make similar predictions when tested on an unseen test set of 500 novel nouns marked with the correct gender. The correlation of predictions of two architectures is 0.67 (\textit{p<0.01}) when tested with correct gender. The accuracy of both models drop significantly when tested in no-gender and neuter gender test sets, which aligns with RNNs oversensitivity to gender \cite{mccurdy2020conditioning}. Both models produce a higher number of inflections that are different from the plural markers when the test set is not marked with gender information. While this positively indicates that the models learn some associations between a noun's gender and its inflected form, it also shows that the models' weakness in making inference with partial information and the predicted inflections may depend too much on the grammatical gender.

When evaluated with the second method based on \citeA{mccurdy2020inflecting}, neither model is found to be correlated strongly to human behavior across any condition, as shown in Table \ref{fig:german-correlations}. When the performance is measured across suffixes (Table \ref{fig:german-correlations}, left), both models perform best ($F1 >= 0.90$) with the /-(e)n/ suffix, which has the highest type frequency in the German corpus, but perform relatively poorly on other suffixes. Despite the lack of a strong and statistically significant correlation between the two model's predictions, the similar trend in the precision, recall, and F1 scores of the models across suffixes suggest that the Transformer performs similarly to the RNN and also struggles to generalize productive rules in minority classes.

\begin{table}[h]
\centering
\resizebox{240pt}{!}{
\begin{tabular}{l|ccc|ccc|ccc}
 & \multicolumn{3}{c|}{\textbf{RNN}} & \multicolumn{3}{c|}{\textbf{Transformer}} & \multicolumn{3}{c}{\textbf{Correlations}} \\ \hline
 & \textbf{Pr.} & \textbf{Rec.} & \textbf{F1} & \textbf{Pr.} & \textbf{Rec.} & \textbf{F1} & \textbf{R - T} & \textbf{R - H} & \textbf{T- H} \\ \hline \hline
\textbf{/-(e)n/} & .91 & .90 & .90 & .94 & .91 & .93 & -.25 & -.07 & .23 \\
\textbf{/-e/} & .79 & .74 & .74 & .74 & .90 & .81 & .41 & .00 & .05 \\
\textbf{/-Ø/} & 1.00 & .61 & .74 & 1.00 & .42 & .57 & .00 & .00 & .00 \\
\textbf{/-s/} & .40 & .29 & .30 & .71 & .21 & .31 & .32 & .44 & .01 \\
\textbf{/-r/} & .51 & .48 & .45 & .70 & .39 & .48 & \textbf{.55} & .22 & .44 \\
\textbf{other} & .00 & .00 & .00 & .00 & .00 & .00 & .00 & -.31 & .00
\end{tabular}
}
\caption{Model performance (precision, recall, F1) on each suffix in CELEX test set, averaged across 25 splits (left). Spearman correlation between models and human data (RNN-Transformer, RNN-Human, Transformer-Human) when tested on the experiment data set (right). Only one of the correlations were statistically significant (\textit{p<0.01}), marked in bold. The Transformer did not predict any /-(e)n/ when tested on this test set.}
\label{fig:german-correlations}
\end{table}


\begin{table*}[t]
\centering
\begin{tabular}{lcccc}
\textbf{Dataset} & \textbf{RNN} & \textbf{Transformer} & \textbf{Correlation} & \textbf{Similarity} \\ \hline\hline
\textbf{All Verbs} & 88.5 & 82.1 & 0.855 & 93.0 \\
\textbf{1st and 2nd Conjugation} & 82.2 & 83.1 & 0.848 & 93.0 \\
\textbf{1st Conjugation} & 84.8 & 86.4 & 0.879 & 96.0 \\
\textbf{2nd Conjugation} & 83.9 & 84.1 & 0.888 & 90.6 \\
\textbf{Dental-stem 2nd Conj} & 83.8 & 84.6 & 0.844 & 93.7
\end{tabular}
\caption{Model comparison on Russian verbs. The reported numbers are accuracy scores on the unseen test set within the dataset per model, and finally the Spearman correlation prediction similarities between the predictions made by the models. All correlation scores have \textit{p<0.01}.}
\label{fig:russian-results}
\end{table*}
\begin{table}[h!]
\centering
\resizebox{240pt}{!}{
\begin{tabular}{llll}
\textbf{English} &\textbf{Russian Inf.} & \textbf{RNN} & \textbf{Transformer} \\ \hline\hline
to protest & \foreignlanguage{russian}{бузить} & \foreignlanguage{russian}{бужу’} &\foreignlanguage{russian}{бу’жу} \\ \hline
to yell & \foreignlanguage{russian}{голосить} & \foreignlanguage{russian}{гололошу'} &\foreignlanguage{russian}{голоси'ю} \\ \hline
to dream & \foreignlanguage{russian}{грезить} & \foreignlanguage{russian}{грежу'} &\foreignlanguage{russian}{гре'жу} \\ \hline
to itch & \foreignlanguage{russian}{зудеть} & \foreignlanguage{russian}{зуде'ю} &\foreignlanguage{russian}{зуде'ю} \\ \hline
\end{tabular}
}
\caption{Examples from the predictions produced by the RNN and the Transformer models on the set of gapped verbs.}
\label{table:russian-samples}
\end{table}

\subsection{The Russian Morphological Gap}

\subsubsection{Method} In order to test the models' behavior on the gaps observed in Russian, we train and evaluate multiple separate instances of both models on 5 different datasets consisting of pairs of non-gapped Russian verbs in present infinitive and their present first person singular form with Cyrillic alphabet.\footnote{Training and testing word forms are obtained from the Russian Dictionary dataset: https://github.com/Badestrand/russian-dictionary} Each training dataset consists of all or a subset of the Russian verbs and their inflected forms, selected based on the criteria listed on Table \ref{fig:russian-results}. We use a 80\% training, 10\% and 10\% testing split for each dataset.\footnote{Training steps for the Transformer are increased proportionally to the training dataset size.} During testing for each dataset, we wug-test the models on the corresponding 10\% test set, as well as a special test set that consists of the gapped verbs.\footnote{We use the list of gapped words provided by \citeA{pertsova2016transderivational}.} 

Since it is technically infeasible to provide the neural sequence-to-sequence models with an explicit option to reject inflecting a given word other than using the models' implicit ability to return the original form, in order to assess whether the models have any difficulty in producing inflected forms resembling the cognitive block humans experience, we compare the models' prediction confidence across the test sets by measuring the likelihood of the predicted inflected forms. If the models struggle to produce inflected forms for the gapped verbs, we would expect to see a  statistically significant drop in the models' prediction confidence for those verbs.

\subsection{Results and Discussion}
The results of the Russian evaluation are shown in Tables \ref{fig:russian-results} and \ref{table:russian-samples}. Regardless of the verb data used to train the models, both neural architectures achieve similar accuracy scores. The architectures make similar predictions, as indicated by statistically significant correlations and high predicted inflection similarities between the predictions made by the two models across all conditions. Examples of the gapped verbs and corresponding inflected forms produced by the neural models are presented in Table \ref{table:russian-samples}. We find no significant change in the models' prediction confidences across the regular and gap test sets. When trained on the `all verbs' dataset, the change in prediction confidences across the test sets is less than 10\% for both models, a decrease from $0.95$ $(\sigma=0.12)$ to $0.86$ $(\sigma=0.18)$ for the RNN model, and a decrease from $0.46$ $(\sigma=0.11)$ to $0.39$ $(\sigma=0.13)$ for the Transformer model, which suggests that neither model captures the gaps in productivity. 

\section{Conclusion \& General Discussion}
We evaluated the Transformer architecture as a model for morphological rule learning and compared it to encoder-decoder RNNs. In our evaluations, we have used three telltale phenomena in morphology that show how children and native speakers expect certain structure in language that cannot be extracted from data per se. Specifically, we have evaluated models on the English past tense where the productive morphological rule is the majority form \cite{berko1958child,albright2003rules,lignos2018morphology}, the German plural where multiple non-majority morphological rules are applied productively \cite{marcus1995german,zaretsky2015no,mccurdy2020inflecting}, and finally the gaps observed in 63 Russian dental-stem second conjugation verbs for which native speakers cannot comfortably produce a first-person-singular present form \cite{sims2006minding,pertsova2016transderivational}. 

In our evaluations on the English past tense, we observed that the Transformer architecture, perhaps due to having more power in capturing distributional properties, is more susceptible to analogical errors when tested on irregular verbs. Our analyses on German plurals suggest that the Transformer, similar to the RNN, is sensitive to the grammatical gender of nouns, is biased towards the most majority suffixes, and does not correlate with human behavior. When evaluated on the gapped Russian verbs, both neural architectures produce inflected first-person-singular forms that are not produced by native speakers, suggesting that the models extract patterns from data and apply them to novel examples without employing the necessary linguistic priors. 

As evinced by the models' predictions across the English past tense, the German plurals, and the Russian gaps, a fundamental challenge for neural architectures such as Transformers and RNNs as models of language learning is the need for additional structural priors that can enable learning categorical productive rules. Incorporating such priors into the neural architectures may not only make them more suitable for cognitive modeling, but also strengthen their generalization capabilities. 

In sum, we did not find any evidence that would suggest Transformers are better cognitive models of morphological rule learning than RNNs. Both neural architectures recapitulate the statistical distribution of inflection alternatives in the training data, and hence are susceptible to similar analogical and distributional errors, suggesting that these models do not seek the necessary linguistic structures sought by humans during morphology acquisition. We also suggest that morphological gaps can be used for measuring the effectiveness of neural architectures as cognitive models in the language domain.

\section{Acknowledgements}
I would like to thank Prof. Charles Yang for his guidance throughout this research and during my studies at the University of Pennsylvania. I would also like to thank Dr. Asli Rehber Beser for her support and reviews, Anton Relin for assisting with the Russian language, and the anonymous reviewers for valuable suggestions. 

\bibliographystyle{apacite}

\setlength{\bibleftmargin}{.125in}
\setlength{\bibindent}{-\bibleftmargin}

\bibliography{references}

\end{document}